 \documentclass[smallabstract,smallcaptions]{dccpaper}

\usepackage{epsfig}
\usepackage{citesort}
\usepackage{amsmath}
\usepackage{amssymb}
\usepackage{color}
\usepackage{url}
\usepackage{multirow}
\usepackage{makecell}

\usepackage{graphicx}
\usepackage{float}
\usepackage{subfig}

\newlength{\figurewidth}
\newlength{\smallfigurewidth}

\begin{document}

\title
{\large
\textbf{Octree-based Learned Point Cloud Geometry Compression: A Lossy Perspective}
}

\author{%
Kaiyu Zheng$^{\ast}$, Wei Gao$^{\ast}$, and Huiming Zheng$^{\ast}$\\[0.5em]
{\small\begin{minipage}{\linewidth}\begin{center}
\begin{tabular}{c}
$^{\ast}$School of Electronic and Computer Engineering, Shenzhen Graduate School, \\
Peking University, Shenzhen, China \\
\url{ { kaiyu_zheng, hmzheng } @stu.pku.edu.cn}, and \url{gaowei262@pku.edu.cn}
\end{tabular}
\end{center}\end{minipage}}
}

\maketitle
\thispagestyle{empty}

\begin{abstract}

Octree-based context learning has recently become a leading method in point cloud compression. However, its potential on lossy compression remains undiscovered. The traditional lossy compression paradigm using lossless octree representation with quantization step adjustment may result in severe distortions due to massive missing points in quantization. Therefore, we analyze data characteristics of different point clouds and propose lossy approaches specifically. For object point clouds that suffer from quantization step adjustment, we propose a new leaf nodes lossy compression method, which achieves lossy compression by performing bit-wise coding and binary prediction on leaf nodes. For LiDAR point clouds, we explore variable rate approaches and propose a simple but effective rate control method. Experimental results demonstrate that the proposed leaf nodes lossy compression method significantly outperforms the previous octree-based method on object point clouds, and the proposed rate control method achieves about 1\% bit error without finetuning on LiDAR point clouds.

\end{abstract}







\Section{Introduction}

Point cloud is a fundamental data structure for presenting 3D scenes. It is widely utilized in different fields, such as visual reality, smart city, and autonomous driving. Since point clouds contain massive numbers of discrete and unordered points, storing and transmitting such large-scale complicated data has become a crucial challenge. Therefore an efficient point cloud compression (PCC) technique is necessary for developing practical point cloud applications. Several schemes are proposed to compress the point cloud effectively. Traditional PCC method MPEG point cloud geometry compression standard (G-PCC) \cite{graziosi2020overview} applies a context arithmetic encoder to predict the node with handcrafted rules. Recently, deep learning-based entropy codecs \cite{fu2022octattention}\cite{wang2021multiscale} have shown outperforming rate-distortion performance compared to handcrafted ones. 

Octree-based context learning is a leading deep learning-based point cloud compression method. It constructs octrees from raw point clouds and transforms point cloud compression into octree compression. A context model is introduced to predict the occupancy codes of octree nodes based on available context. Recently, with the utilization of self-attention \cite{fu2022octattention}, octree-based methods show outstanding performance. 

Though octree-based methods present promising potential for the future, the octree structure compression is fundamentally lossless. Previous octree-based lossy compression is achieved by quantization step adjustment when constructing the octree, which may result in severe distortions due to massive missing points in quantization. As lossy compression is crucial in multimedia research to balance file size reduction with acceptable quality degradation, the lossy compression on octree is worthwhile further exploring.

In this paper, we first analyze the effect of quantization step adjustment on different point clouds. For object point clouds that are not suitable for quantization step adjustment, we  propose a new octree-based lossy compression approach that decomposes the octree into non-leaf nodes and leaf nodes, and performs lossy compression only on leaf nodes by bit-wise coding and binary value prediction. Moreover, we discover a variable rate approach for LiDAR point clouds, and present the first octree-based rate control method for LiDAR point clouds.

The contribution of our work can be summarized as:

\begin{itemize}
    \item We systematically research octree-based lossy compression. We analyze the data characteristics of different point clouds and propose that quantization step adjustment is inappropriate for object point clouds lossy compression.
    \item We propose a new octree-based lossy compression method for object point clouds that achieves lossy compression by performing bit-wise coding and binary prediction on leaf nodes. To the best of our knowledge, this is the first octree-based lossy method without quantization step adjustment.
    \item We further explore the variable rate approach for LiDAR point clouds and propose a simple but effective rate control method. Experiments prove that our method can achieve about 1\% bit error without finetuning.
\end{itemize}

\Section{Related Work}

\SubSection{Octree-based Point Cloud Geometry Compression}

Octree-based methods construct an octree from the point clouds and convert point cloud compression into tree structure compression. The MPEG point cloud geometry compression standard (G-PCC) compresses the point cloud through the octree structure by predicting the occupancy code with handcrafted rules \cite{graziosi2020overview}. Recently, neural networks are utilized for octree compression and presenting superior performance. OctSqueeze \cite{huang2020octsqueeze} initially proposes an MLP-based context model to exploit the correlation among nodes with ancestor features, but the receptive field is limited. Therefore, OctAttention \cite{fu2022octattention} introduces self-attention with a large context window, with the auto-regressive structure, both ancestor and sibling features are able to be exploited. Nevertheless, the auto-regressive structure brings intolerable decoding time, thus OctFormer \cite{cui2023octformer} and EHEM \cite{song2023efficient}  are proposed to tackle this problem by replacing the auto-regressive structure with non-overlapping context window. However, these methods achieve lossy compression by quantization step adjustment, there is still few researches on octree-based lossy compression.

\SubSection{Point Cloud Geometry Lossy Compression}

Previous point cloud lossy compression approaches are mainly voxel-based learned methods, which quantize the raw point cloud into voxel grids and classify the voxel occupancy via the neural network. Earlier works are based on uniform voxels \cite{wang2021lossy}, where 3D dense convolutional layers are utilized to exploit the spatial correlations and cost excessive computational complexity in both time and space dimension. Therefore, sparse 3D convolution is introduced to overcome the complexity. In SparsePCGC \cite{wang2022sparse} and PCGCv2 \cite{wang2021multiscale}, auto-encoders with 3D sparse convolution present noticeable compression performance. Recent works also exploit approaches of variable rates \cite{qi2024variable}\cite{yu2024dynamic} by considering latent representation and computational complexity. However, octree-based lossy compression is still understudied.

\begin{figure}
    \centering
    \subfloat[]{
        \includegraphics[width=0.46\linewidth]{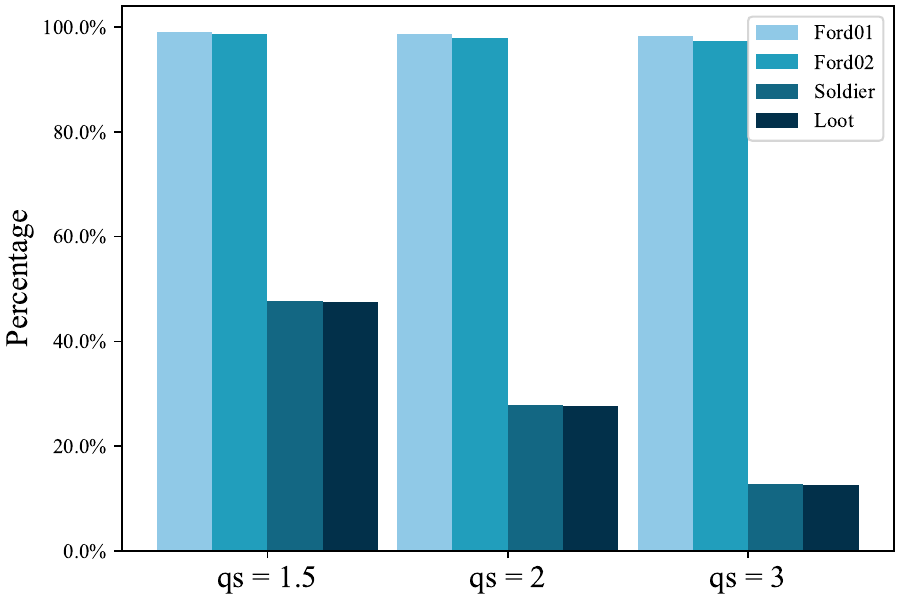}
    }
    \subfloat[]{
        \includegraphics[width=0.23\linewidth]{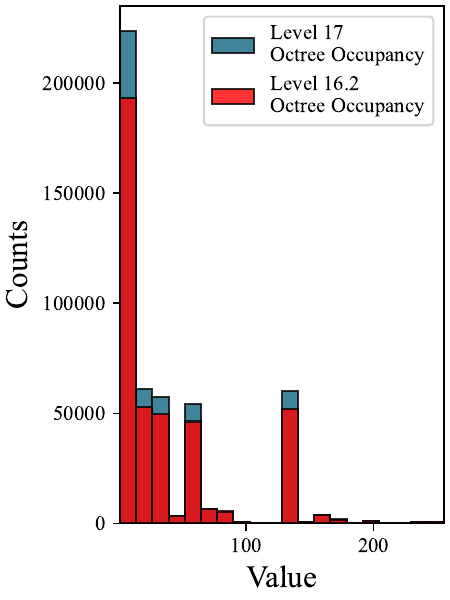}
    }
    \subfloat[]{
        \includegraphics[width=0.23\linewidth]{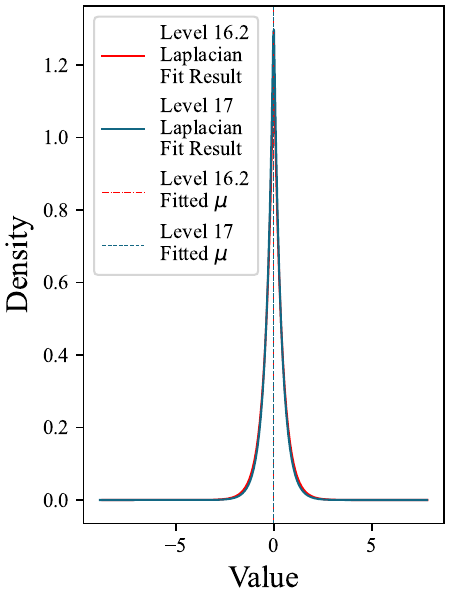}
    }
    \caption{Data Analysis. (a) The ratio of remaining points when setting different $qs$. (b) Histogram of octree occupancy distribution in depth level 16.2 and 17. (c) Laplacian fitting results of the 
 context model output in depth level 16.2 and 17.}
    \label{fig:analysis}
\end{figure}

\Section{Proposed Method}

\SubSection{Analysis of the Effects of Quantization Step Adjustment}

Previous octree-based PCC methods are essentially octree structure lossless compression methods, the lossy compression can only be achieved by quantization step $qs$ adjustment. Denote the original point cloud as $PC$, $qs$ with given depth level $L$ is 

\begin{equation}
    qs_L = \frac{\mathrm{max}(PC)-\mathrm{min}(PC)}{2^L-1}. \label{qs_calculate}
\end{equation}

Altering the $qs$ results in neighboring points being rounded to the same integer point and causes point number loss. As the data characteristics vary differently between object point cloud and LiDAR point cloud, we set up different $qs$ to explore the effect on these two types of data. We select \textit{Soldier} and \textit{Loot} sequences from the object dataset MPEG 8i \cite{d20178i} and \textit{Ford01} and \textit{Ford02} sequences from the LiDAR dataset Ford \cite{pandey2011ford}. We set $qs=\{1.5,2,3\}$ and calculate the ratio of remaining points to original points, as shown in Fig. \ref{fig:analysis}(a).

As can be seen, in object point clouds, quantization step adjustment causes a large number of points to be rounded to the same integer point. Consequently, the number of points in reconstructed point clouds drops significantly compared to original point clouds, leading to severe distortion as merged points can not find corresponding original points. In contrast, adjusting the quantization step for LiDAR points clouds can effectively bring sparsely distributed points closer and enrich the context, while the point number barely drops.

Hence, the quantization step adjustment is appropriate for LiDAR point clouds but not object point clouds, therefore we propose a new octree-based PCC method for object point clouds. We find that leaf nodes make up more than 70\% of the entire octree in object point clouds. Thus, performing lossy compression only on leaf nodes can effectively reduce the bitrate. For LiDAR point clouds, we discover approaches for variable rate and rate control. These methods will be introduced below.

\begin{figure}
    \centering
    \includegraphics[width=0.98\linewidth]{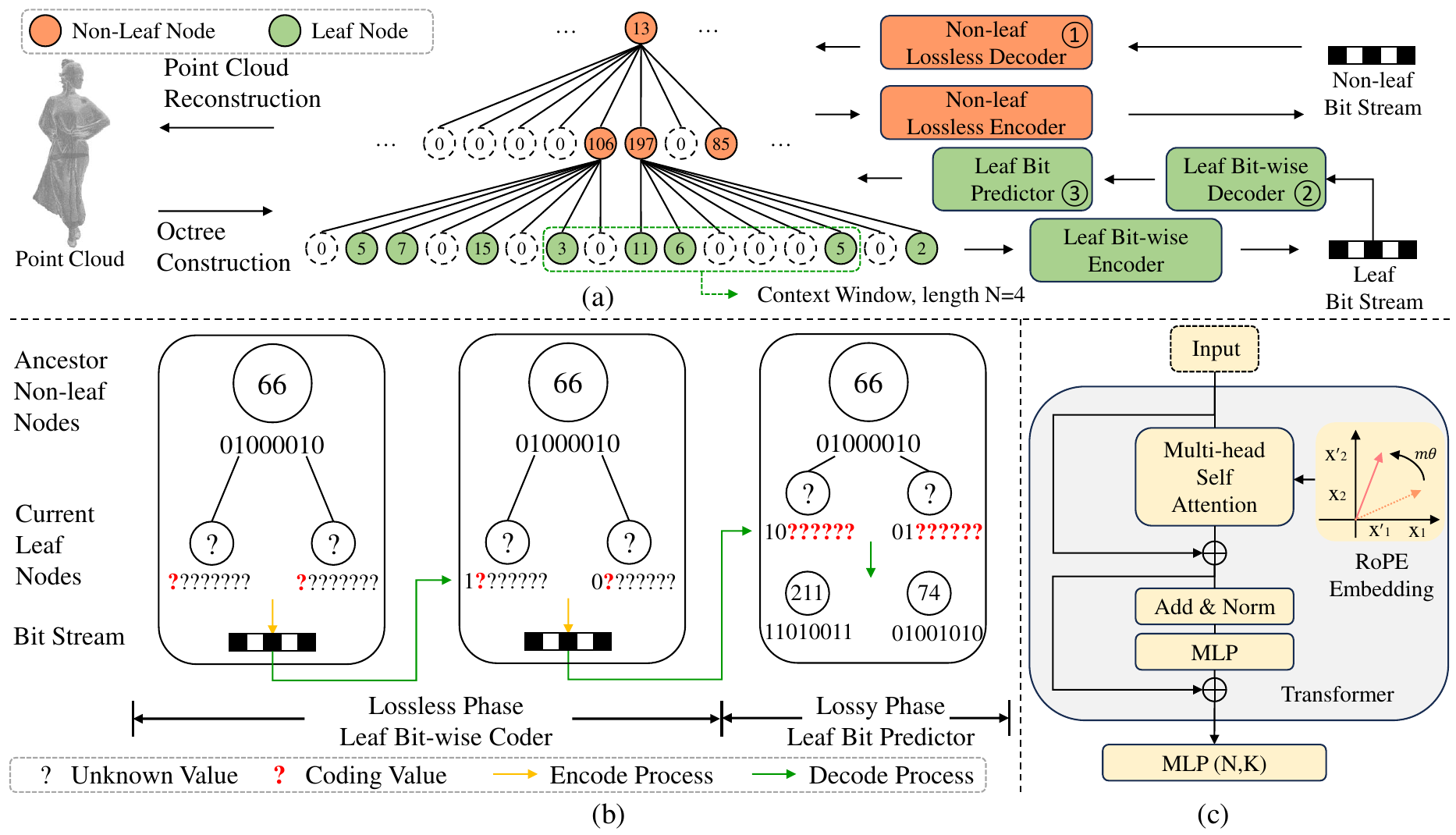}
    \caption{Proposed leaf nodes lossy compression framework. (a) The process of compression, the number in the octree represents occupancy, and the number in the coding module represents the decoding order. (b) Details of leaf nodes compression, leaf lossless steps $s$ is set to 2 in the figure. (c) Structure of backbone context model. For non-leaf coder, $K$ is set to 255. For leaf bit-wise coder, $K$ is set to 2. For leaf bit predictor, $K$ is set to 8.}
    \label{fig:leaf}
\end{figure}

\SubSection{Leaf Nodes Lossy Compression on Object Point Clouds}

We propose a new octree-based lossy compression by performing bit-wise coding and binary value prediction on leaf nodes for object point clouds. We decompose the occupancy code into 8-bit binary values, i.e., 8 child nodes, which denote the occupancy of the eight sub-cubes that belong to the leaf node. We only transmit part of binary values, while the rest values consume no bit stream, their values are predicted on the decoder side. We only perform lossy compression on leaf nodes because the non-leaf nodes correspond to large volume cubes in voxel space, if a non-leaf node is distorted, it causes a wide range of occupancy errors in voxel space. The leaf node lossy method constrains distortion to a few incorrectly predicted leaf binary values and the massive missing points are avoided, thus maintaining the reconstruction quality.

\noindent\textbf{Leaf Lossless Phase:} For the leaf lossless phase, we propose a bit-wise auto-regressive lossless compression. In each step, all leaf nodes' one binary value in the context window are parallelly coded. The auto-regressive process exploits the dependency among child nodes, thus utilizing the former coded child nodes to assist with the later child nodes' prediction and save the bitrate.

For the $i$-th octree node in a context window, denote the features of ancestor non-leaf nodes as $C_a$, the binary values of an octree node $x_i$ as $b_{i,1}, \dots , b_{i,8}$, the leaf bit-wise context model as $\theta_1$. The probability of the $j$-th binary value is 

\begin{equation}
    P(b_{i,j})=P_{\theta_1}(b_{i,j}|C_a,\{b_{p,q}|1 \le p \le i, 1 \le q \le j-1\} ) . 
\end{equation}

Note that constrained by the auto-regressive structure, it is only possible to perform parallel encoding for all binary values. When decoding, in order to ensure the causal sequence, the later binary values have to wait for the decoding results of former binary values to keep the context exactly the same as the encoder, thus the decoding process is serial. The rate-distortion control can be utilized by altering the leaf lossless steps $s$, and the $s$ ranges from 0 to 8. Higher $s$ leads to higher bitrate, better reconstruction performance and longer decoding time. 


\noindent\textbf{Leaf Lossy Phase:} The leaf lossy phase does not consume bitrate and is only performed on the decoder side. The leaf lossy phase is essentially a multi-label prediction task for a leaf bit predictor model $\theta_2$, aiming at predicting the binary value of all untransmitted binary values at one time based on ancestor and sibling contexts. The probability of the $j$-th to $8$-th binary value of $i$-th octree node in a context window is

\begin{equation}
   P(b_{i,j},\dots,b_{i,8})=P_\theta(b_{i,j},\dots,b_{i,8}|C_a,\{b_{p,q}|1 \le p \le i, 1 \le q \le j-1\} ) . 
\end{equation}

Different from the cross entropy loss in lossless compression tasks, the loss function in the leaf lossy phase is the mutation of binary cross entropy loss \cite{liu2021emerging}, the loss of $i$-th node is denoted as: 
 
\begin{align}
    Loss_i=-\frac{1}{8-s} \sum_{j}^{8-s} \left [    y_j\cdot \log ((1+\exp(-b_j ))^{-1}) + (1-y_j)\cdot \log \left (\frac{\exp(-b_j )}{(1+\exp(-b_j ))} \right ) \right ],
\end{align}
where $y$ denotes the ground truth binary value. The index $j\in \{1,\dots,8-s\}$ and $y_j\in \{0,1\}$. 

\SubSection{Variable-Rate Lossy Compression on LiDAR Point Clouds with Rate Control}

Due to the sparseness, the aforementioned leaf nodes lossy compression is not appropriate for LiDAR point clouds with insufficient context. Meanwhile, previous octree-based PCC methods \cite{fu2022octattention}\cite{song2023efficient} have shown surpassing compression performance on LiDAR data. However, these methods only achieve lossy compression by truncating the octree at varying depths, which limits the available bitrates and makes it hard to reach specific bitrates needed for bandwidth constraints. Hence we further explore the variable-rate approach by analyzing more refined quantization steps in deep depths. 

When truncating the octree, depth $L$ in Equation \ref{qs_calculate} is set to integer. Denote two neighboring truncating depths as $L$ and $L+1$, the truncated octrees are constructed with $qs_L$ and $qs_{L+1}$. Denote the bit per point (bpp) for these octrees as truncating points $bpp_L$ and $bpp_{L+1}$, the target of the variable rate approach is to obtain $bpp$ between $bpp_L$ and $bpp_{L+1}$. We manually set the quantization step $qs_{\hat{L}}$ between $qs_L$ and $qs_{L+1}$. The actual octree depth constructed from $qs_{\hat{L}}$ is $L+1$, and we analyze the data distribution similarity between the octrees constructed from $qs_{\hat{L}}$ and $qs_{L+1}$. We set $\hat{L}=16.2$ and present the occupancy distribution and the Laplacian fitting result of the feature output from the same coding context model, as shown in Fig. \ref{fig:analysis}(b)(c). It can be seen that the input occupancy distribution tends to be similar between octrees from $qs_{16.2}$ and $qs_{17}$, only differs in the total number of octree nodes. Meanwhile, the Laplacian fitting result of the output feature also demonstrates that the coding context model generalizes well on octrees constructed from different $qs$ with the same depth. Based on these analyses, by altering the $qs_{\hat{L}}$, variable rates between $bpp_L$ and $bpp_{L+1}$ can be achieved without finetuning the coding context model. 

Denote the bit per octree node cost as $c$, the total number of points as $N_{PC}$, the total number of octree nodes as $N_{Oct}$, and the relation between bit per point (bpp) and $c$ is $bpp \cdot N_{PC}=c\cdot N_{Oct}$. In practice, the cost $c$ is nearly the same among octrees with the same depth. As $N_{Oct} \propto -qs$, and $N_{PC}$ is constant, we empirically propose that $bpp$ and $qs$ are linearly related. Hence, we propose a simple but effective LiDAR point cloud rate control approach. Given the target bpp $bpp_t$ with trained context model, first obtain pre-calculated $bpp_L$ and $bpp_{L+1}$ with their $qs_L$ and $qs_{L+1}$, where $bpp_L \le bpp_t \le bpp_{L+1}$. The $qs_{\hat{t}}$ that constructs the octree to achieve $bpp_t$ is estimated as 

\begin{equation}
    qs_{\hat{t}}=\frac{bpp_{t}-bpp_{L}}{bpp_{L+1}-bpp_{L}}(qs_{L+1}-qs_{L})+qs_{L} . \label{target qs}
\end{equation}

We achieve the rate control by utilizing the linear equation, experiments have shown the effectiveness of our approach. Needed to mention, this approach is applicable to all octree-based PCC methods that present data distribution similarity in the same depth, making it convenient and universal.

\SubSection{Backbone Context Model}

We follow the checkerboard-like coding strategy in \cite{song2023efficient}. The structure of the backbone context model is shown in Fig. \ref{fig:leaf}(c). We utilize the occupancy, level, and octant index from $D$ levels of ancestors and sibling nodes as context features. These features are embedded and processed by the transformer network and eventually output the probability of occupancy or binary value. Note that only the indexes of lossy phase binary values are supervised by $Loss_i$ when training. In the inference stage, the binary values of lossy phase is determined by the Sigmoid function.

As the octree sequence only consists of occupied cubes, the actual distance between two neighboring nodes in voxel space is indeterminate. In the context model, we introduce the Rotary Position Embedding scheme (RoPE) \cite{su2024roformer} into the transformer. It enhances the transformer by rotating the embeddings in a multi-dimensional space based on their positions, allowing the model to maintain a consistent representation of relative positions while incorporating absolution positions.



\begin{table}[!tbp]
\centering
\renewcommand\arraystretch{1}
\caption{Performances of BD-Rate gain compared with other methods.}
\resizebox{\linewidth}{!}{
\begin{tabular}{c|cccccccc}
\hline

\hline
    \multirow{2}{*}{Data}&\multicolumn{2}{c}{G-PCC} &\multicolumn{2}{c}{OctAttention}&\multicolumn{2}{c}{EHEM}&\multicolumn{2}{c}{PCGCv2}\\ 
\cline{2-9}
&D1&D2&D1 &D2&D1 &D2&D1&D2\\
\hline
Redandblack &-49.65&-66.65&-75.85&-60.60&-71.68&-58.63&-5.38&-19.02\\
Loot&-33.89&-62.20&-87.59&-51.10&-78.68&-39.69&-6.40&-15.43\\
Phil&-55.97&-70.49&-66.37&-74.57&-70.51&-75.64&-13.74&-52.90\\
Ricardo&-57.16&-68.24&-68.20&-40.33&-72.02&-48.33&-16.82&-72.96\\
\textbf{Avg}&\textbf{-49.17}&\textbf{-67.90}&\textbf{-74.50}&\textbf{-56.65}&\textbf{-73.22}&\textbf{-55.57}&\textbf{-10.59}&\textbf{-40.08}\\
\hline
\end{tabular}
}
\label{table:psnr compare}
\end{table}

\begin{figure}
    \centering
    \subfloat{
    \includegraphics[width=0.23\linewidth]{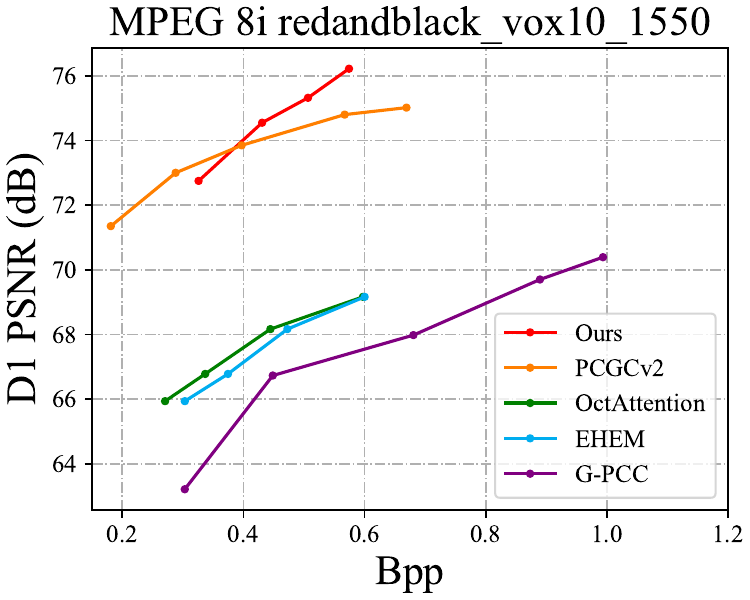}
    }
    \subfloat{
    \includegraphics[width=0.23\linewidth]{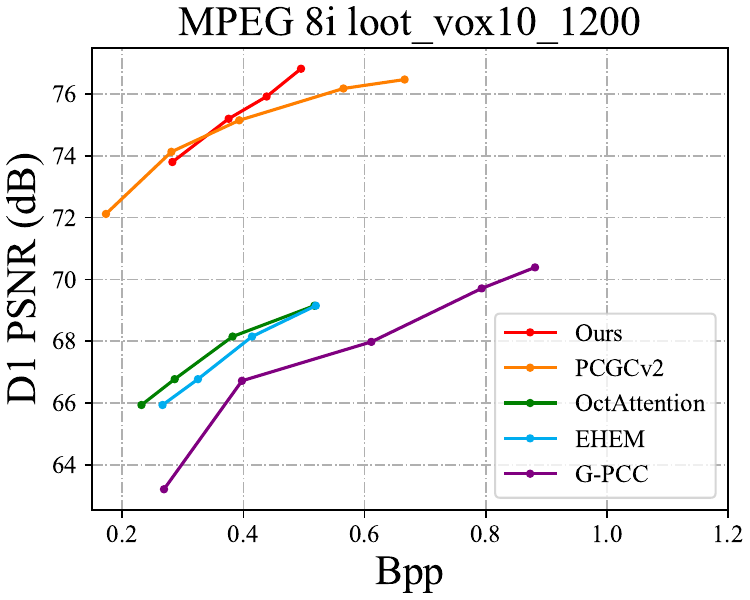}
    }
    \subfloat{
    \includegraphics[width=0.23\linewidth]{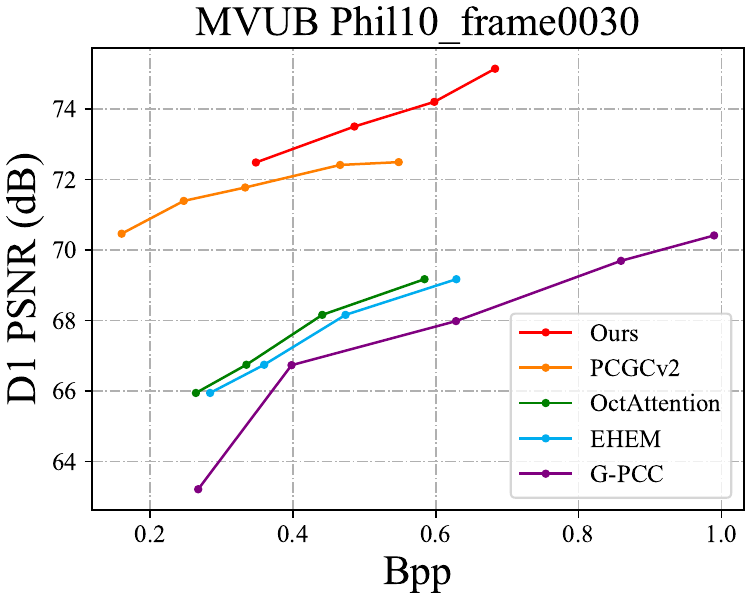}
    }
    \subfloat{
    \includegraphics[width=0.23\linewidth]{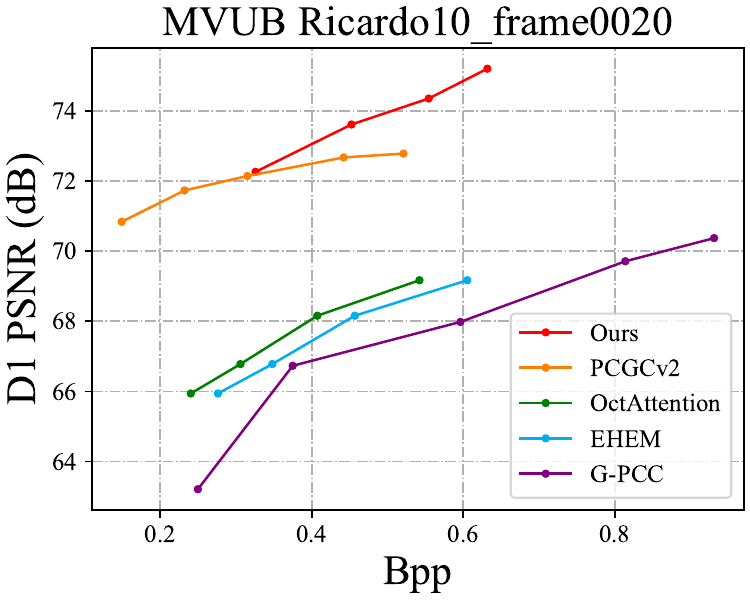}
    }
    \caption{Performance comparison using rate-distortion curve on MPEG 8i and MVUB dataset.}
    \label{fig:rd_curve_lossy}
\end{figure}

\Section{Experiments}

\SubSection{Experimental Settings}

We experiment on LiDAR and object datasets to evaluate our lossy compression methods. Following the setting in \cite{fu2022octattention}, we experiment on the 8i Voxelized Full Bodies (MPEG 8i) dataset \cite{d20178i} and the Microsoft Voxelized Upper Bodies (MVUB) dataset \cite{loop2016microsoft}. We utilize \textit{Andrew10, David10, Sarah10} sequences in MVUB, \textit{Soldier10 and Longdress10} sequences in MPEG 8i for training. For LiDAR point cloud, we adopt the Ford dataset \cite{pandey2011ford} and follow the dataset setting in \cite{song2023efficient}, where sequence 01 is utilized as the train dataset and sequence 02 and 03 are utilized as the test dataset. Meanwhile, we normalize the raw point cloud data into $[-1, 1]^3$ with the octree depth from 15 to 17. We utilize the point-to-point (D1 PSNR), point-to-plane (D2 PSNR), chamfer distance (CD) and BD-Rate \cite{ctc} to measure the reconstruction quality in lossy compression. The bitrate is measured by bpp. The bit error $E$ is estimated as $E=\frac{\mathrm{abs} (R_t-R_a)}{R_t}$, where $R_t$ and $R_a$ represent target bitrate and actual bitrate. 

We compare our method with octree-based methods EHEM \cite{song2023efficient}, OctAttention \cite{fu2022octattention}, voxel-based method PCGCv2 \cite{wang2021multiscale} and traditional method MPEG G-PCC (TMC13v14, generally used baseline \cite{wang2022sparse}) \cite{tmc13} on object point clouds. For LiDAR point clouds, we implement our method on both our context model with 8 transformer layers and OctAttention \cite{fu2022octattention} to evaluate the effectiveness and generalization, and the target rate is set from low bitrate $r_1$ to high bitrate $r_5$, i.e., $\{12.5, 13, 14, 15, 16\}$.

All training and evaluation experiments are implemented on the NVIDIA GeForce RTX 3090 GPU. The learning rate is set to $2\times 10^{-4}$ and epochs is set to 10. The context window length is set to 2048 in the leaf bit predictor model and 1024 for other models. When compared to the previous octree-based methods on the object point cloud, the lossy compression is achieved by the quantization step adjustment, where $qs \in \{1.2, 1.4, 1.6, 1.8\}$. The leaf lossless steps $s$ is set from 1 to 4.

\SubSection{Experimental Results}

\noindent\textbf{Object Point Cloud Lossy Compression: }The quantitative results of the rate-distortion performance on object point clouds are shown in Fig. \ref{fig:rd_curve_lossy} and Table \ref{table:psnr compare}. The proposed leaf nodes lossy compression method significantly surpasses previous octree-based methods in object point cloud lossy compression. Meanwhile, our method outperforms the voxel-based method PCGCv2 with better reconstruction performance, especially at higher bitrates. This demonstrates that our method is the first octree-based method to outperform the widely-used voxel-based method PCGCv2.

The inference time are evaluated on sequences \textit{Redandblack} and \textit{Phil}, as shown in Table \ref{table:tiime ablation}. The encoding time of our method is relative to other methods, but the decoding time costs a lot. Due to the causal restriction on the decoder side, the order of non-leaf and leaf nodes and the auto-regressive bit-wise coding are both serial.  

\begin{table}[!tbp]
\centering
\renewcommand\arraystretch{1}
\caption{Bit error of proposed rate control method on different models.}
\begin{tabular}{c|c|cccccc}
\hline

\hline
Model&Data&$r1$&$r2$&$r3$&$r4$&$r5$&\textbf{Avg}\\
\hline

\multirow{2}*{OctAttention}&Ford02&1.71\%&1.73\%&0.17\%&1.32\%&1.37\%&\textbf{1.16\%}\\
&Ford03&0.56\%&1.28\%&1.48\%&0.26\%&1.07\%&\textbf{1.04\%}\\
\hline
\multirow{2}*{Our model}&Ford02&2.14\%&0.13\%&1.42\%&1.30\%&0.79\%&\textbf{1.26\%}\\
&Ford03&0.92\%&1.26\%&0.56\%&0.86\%&1.61\%&\textbf{0.93\%}\\

\hline
\end{tabular}
\label{table:bit error}
\end{table}

\begin{figure*}[!t]
    \centering
    \subfloat{
    \includegraphics[width=0.315\linewidth]{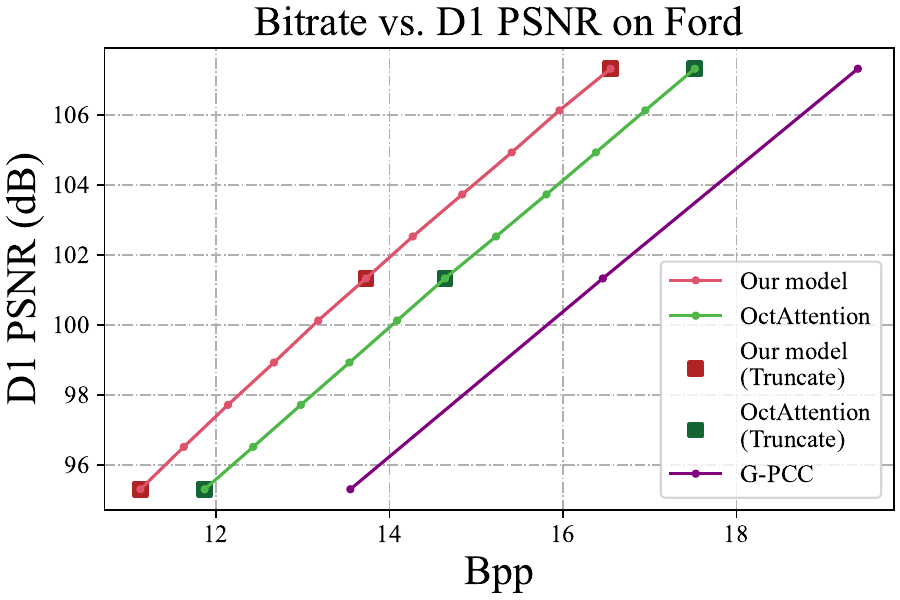}
    }
    \subfloat{
    \includegraphics[width=0.315\linewidth]{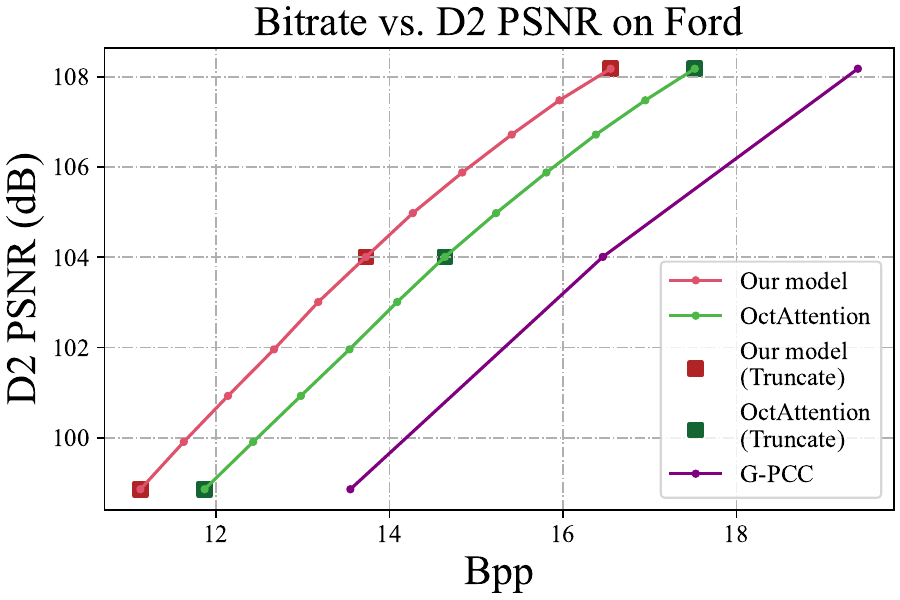}
    }
    \subfloat{
    \includegraphics[width=0.315\linewidth]{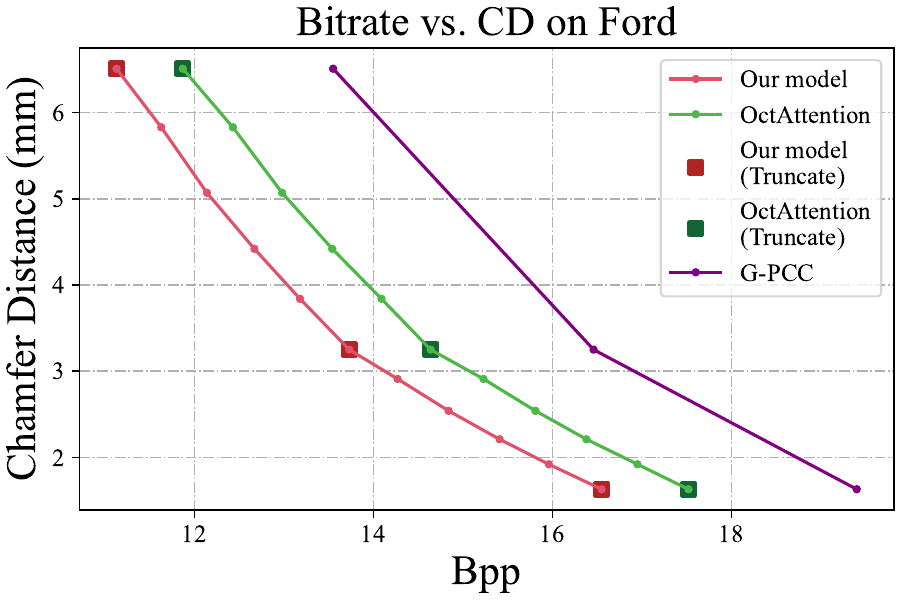}
    }
    \caption{Performance comparison using rate-distortion curve on Ford dataset with variable rate. The square points are results when using the truncating way.}
    \label{fig:rd_curve_ford}
\end{figure*}

\noindent\textbf{LiDAR Point Cloud Variable Rates and Rate Control: }The quantitative results of the rate-distortion performance on LiDAR cloud with variable rates are shown in Fig. \ref{fig:rd_curve_ford}. Results obtained from previous truncating methods are also presented.  Compared to the truncating way, our method of setting $qs_{\hat{L}}$ can achieve variable rates without finetuning, and it suits diverse models. Meanwhile, the rate-distortion curve between two truncating points demonstrates a clear linear relationship, which provides strong support for our rate control method.

The $qs_{\hat{t}}$ is calculated by Equation \ref{target qs} with truncating points to construct the octree for target bitrate $bpp_t$. The bit error of the proposed rate control method is presented in Table \ref{table:bit error}. Our rate control method achieves about 1\% bit error, which is a very acceptable performance as our method needs no finetuning but only evaluation of truncating points. The well-controlled bit errors on different models demonstrate the generalization of the proposed method. Also, we find that the closer the target bitrate is to the truncating points, the smaller the bit error is.

\begin{table}[!tbp]
\centering
\renewcommand\arraystretch{1}
\caption{The ablation study with the encoding time and decoding time. The BD-Rate refers to the gain calculated by the proposed method relative ablation methods.}
\resizebox{\linewidth}{!}{
\begin{tabular}{c|cccc|cccc}
\hline

\hline
&\multicolumn{2}{c}{\makecell[c]{Ours\\ w/o leaf\\ w/o RoPE} } &\multicolumn{2}{c|}{\makecell[c]{Ours\\ w/o leaf}}&G-PCC&OctAttention&PCGCv2 &Ours \\ 
\cline{2-9}
Data&\multicolumn{2}{c}{BD-Rate}&\multicolumn{2}{c|}{BD-Rate}&Time(s)&Time(s)&Time(s)&Time(s)\\
&D1&D2&D1 &D2&Enc/Dec&Enc/Dec&Enc/Dec&Enc/Dec \\
\hline
Redandblack &-70.93&-55.62&-69.35&-53.42&1.97/0.47&1.18/302&1.27/0.55&2.93/12.85\\
Phil&-74.11&-80.47&-68.95&-76.45&3.01/0.83&1.30/514&1.35/1.09&3.73/11.28\\
\textbf{Avg}&\textbf{-72.52}&\textbf{-68.05}&\textbf{-69.15}&\textbf{-64.94}&\textbf{2.49/0.65}&\textbf{1.24/408}
&\textbf{1.31/0.82}&\textbf{3.33/12.07}\\
\hline
\end{tabular}
}
\label{table:tiime ablation}
\end{table}

\SubSection{Ablation Study}

We perform ablation study experiments on the object dataset to evaluate the proposed lossy compression method. We construct two models, context model with no leaf nodes lossy compression, denoted as "ours w/o leaf", which achieves lossy compression by quantization step adjustment. Context model with no leaf nodes lossy compression and no RoPE position embedding, denoted as "ours w/o leaf, w/o RoPE". The results are shown in Table \ref{table:tiime ablation}, which demonstrates that the proposed leaf nodes compression significantly improves the lossy compression performance. Meanwhile, "ours w/o leaf" performs better than "ours w/o leaf, w/o RoPE", which demonstrates the effectiveness of the RoPE embedding.

\Section{Conclusion}

In this paper, we first evaluate the impact of quantization step adjustment on different point clouds. For object point clouds where quantization step adjustment is deemed ineffectual, we introduce an innovative octree-based lossy compression method. This method involves decomposing the octree into non-leaf and leaf nodes, followed by the lossy compression solely on the leaf nodes through bit-wise coding and binary value prediction. Additionally, we investigate variable rate methodologies for LiDAR point clouds, and propose a simple but effective approach for rate control. Experiment results prove the effectiveness of the proposed methods. As this paper only carries out preliminary explorations about octree-based lossy compression, the compression performance and speed remain explored in the future.






\Section{References}
\bibliographystyle{IEEEbib}
\bibliography{main}

\end{document}